\documentclass[journal,twoside,web]{ieeecolor}
\usepackage{generic}
\usepackage{cite}
\usepackage{amsmath,amssymb,amsfonts}
\usepackage{graphicx}
\usepackage{textcomp}
\usepackage{algorithmicx}
\usepackage[ruled,vlined]{algorithm2e}
\usepackage{algpseudocode}
\usepackage{amsmath}
\usepackage{amsmath,amssymb,amsfonts}
\usepackage[mathscr]{eucal}
\usepackage{graphicx}
\usepackage{textcomp}
\usepackage{graphicx}
\usepackage{float}
\usepackage{subfig}
\usepackage{url}
\usepackage{longtable}
\usepackage{booktabs}
\usepackage{multirow}
\def\BibTeX{{\rm B\kern-.05em{\sc i\kern-.025em b}\kern-.08em
    T\kern-.1667em\lower.7ex\hbox{E}\kern-.125emX}}
\begin{document}
\title{Automated Mapping the Pathways of Cranial Nerve II, III, V, and VII/VIII: A Multi-Parametric Multi-Stage Diffusion Tractography Atlas}
\author{Lei~Xie, Jiahao~Huang, Jiawei Zhang, Jianzhong~He, Yiang~Pan, Guoqiang~Xie, Mengjun~Li, Qingrun~Zeng, Mingchu Li, Yuanjing~Feng
\thanks{This work was supported in part by the National Natural Science Foundation of China (No. U22A2040, U23A20334, 62403428); Natural Science Foundation of Zhejiang Province (No. LQ23F030017, LMS25F030004); Zhejiang Province Science and Technology Innovation Leading Talent Program (No. 2021R52004). (\textit{Corresponding author: Qingrun~Zeng, Mingchu Li, and Yuanjing~Feng (fyjing@zjut.edu.cn).})}
\thanks{Lei~Xie, Qingrun~Zeng, and Yuanjing~Feng are with the Advanced Interdisciplinary Science and Technology, Zhejiang University of Technology, Hangzhou 310023, China (xielei@zjut.edu.cn).}
\thanks{Jiahao~Huang, Jiawei Zhang, Yiang~Pan and Jianzhong~He, are with the College of Information Engineering, Zhejiang University of Technology, Hangzhou 310023, China.}
\thanks{Mengjun~Li is with Department of Radiology, Xiangya Hospital, Central South University, Changsha 410000, China.}
\thanks{Mingchu Li is with the Department of Neurosurgery, Capital Medical University Xuanwu Hospital, Beijing 100053, China.}
\thanks{Guoqiang~Xie is with the Department of Neurosurgery, Nuclear Industry 215 Hospital of Shaanxi Province, Xianyang 712000, China.}
\thanks{This work focuses on the following tracts: optic nerve (CN II), oculomotor nerve (CN III), trigeminal nerve (CN V), facial-vestibulocochlear nerve (CN VII/VIII). The facial-vestibulocochlear nerve consists of CN VII and CN VIII, which emerge from the lower lateral pons across the cisternal portion and enter the internal auditory canal, so we refer to them collectively as CN VII/VIII.}
}

\maketitle

\begin{abstract}
\textit{Objective:} Cranial nerves (CNs) play a crucial role in various essential functions of the human brain, and mapping their pathways from diffusion MRI (dMRI) provides valuable preoperative insights into the spatial relationships between individual CNs and key tissues. However, mapping a comprehensive and detailed CN atlas is challenging because of the unique anatomical structures of each CN pair and the complexity of the skull base environment. \textit{Method:} In this work, we present what we believe to be the first study to develop a comprehensive diffusion tractography atlas for automated mapping of CN pathways in the human brain. The CN atlas is generated by fiber clustering by using the streamlines generated by multi-parametric fiber tractography for each pair of CNs. Instead of disposable clustering, we explore a new strategy of multi-stage fiber clustering for multiple analysis of approximately 1,000,000 streamlines generated from the 50 subjects from the Human Connectome Project (HCP). \textit{Results:} Quantitative and visual experiments demonstrate that our CN atlas achieves high spatial correspondence with expert manual annotations on multiple acquisition sites, including the HCP dataset, the Multi-shell Diffusion MRI (MDM) dataset and two clinical cases of pituitary adenoma patients. \textit{Conclusion:} The proposed CN atlas can automatically identify 8 fiber bundles associated with 5 pairs of CNs, including the optic nerve CN II, oculomotor nerve CN III, trigeminal nerve CN V and facial-vestibulocochlear nerve CN VII/VIII, and its robustness is demonstrated experimentally. \textit{Significance:} This work contributes to the field of diffusion imaging by facilitating more efficient and automated mapping the pathways of multiple pairs of CNs, thereby enhancing the analysis and understanding of complex brain structures through visualization of their spatial relationships with nearby anatomy.

\end{abstract}

\begin{IEEEkeywords}
Diffusion MRI, Tractography, Cranial nerves, Multi-stage clustering, Brain tumor
\end{IEEEkeywords}

\section{Introduction}
\label{sec:introduction}
\IEEEPARstart{C}{RANIAL} nerves (CNs), which originate from the brain as 12 paired structures, serve critical sensory and motor functions, including auditory, olfactory, visual, gustatory and facial expression-mediated emotional communication~\cite{zolal2016comparison,yoshino2016visualization,muhammad2024intraoperative}. Damage to any CNs, affected by diseases such as trigeminal neuralgia, craniopharyngioma (CP), pituitary adenoma (PA) or brain tumor, can lead to considerable deficits in critical functions~\cite{li2024tractography,behan2017comparison,diakite2025dual}. Identifying CNs allows the visualisation of their spatial relationships with nearby structures, such as tumor or lesions, which is crucial for preoperative diagnosis and treatment planning~\cite{he2021comparison,xie2023cntseg}.

\begin{figure*}[h]
	\centering
	\includegraphics[width=0.75\textwidth]{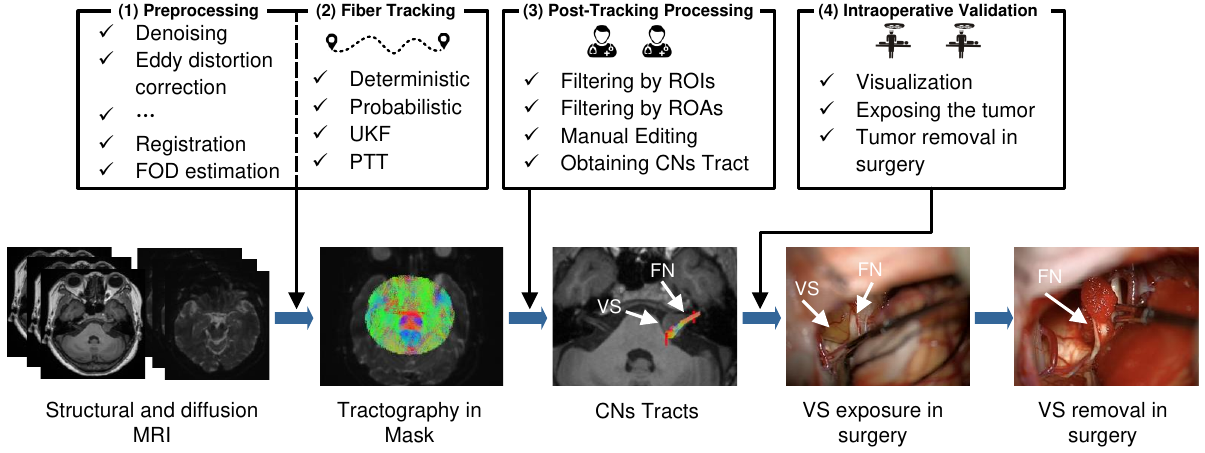}
	\caption{Overview of typical CN tractography pipelines. The raw diffusion-weighted images are often preprocessed to remove eddy current and phase distortion artifacts or address other signal-related issues. The preprocessed images are resolved by FOD estimation models for fiber orientation on each voxel and its associated diffusion metrics, and fiber tracking is performed inside the Mask to generate the tractograms by the chosen fiber tractography method. Then, CNs tracts are finally obtained from the tractograms by post-tracking processing operations, including filtering by ROIs, ROAs, and manual editing. Visualization of CNs tracts can provide the location of key tissues in relation to each other intraoperatively for skull base surgery and better assist the surgeon in removing tumors during surgery.}
	\label{fig:1}
\end{figure*}
Recently, diffusion MRI (dMRI) tractography has been successfully applied to CN identification, offering the advantage of non-invasive in vivo mapping of three-dimensional trajectories~\cite{jacquesson2019overcoming,jacquesson2019full,hu2024preoperative,xie2024anatomy}. 
Early studies employed region-of-interest (ROI) selection strategies to manually extract anatomically relevant tracts from streamlines generated by fiber tractography algorithms, including deterministic, probabilistic~\cite{dyrby2007validation}, Unscented Kalman Filter (UKF)~\cite{malcolm2010filtered}, and parallel transport tractography (PTT)~\cite{aydogan2020parallel}. 
As illustrated in Fig.~\ref{fig:1}, a critical limitation of ROI-based CN identification pipelines stems from their dual reliance on expertise in dMRI tractography and neuroanatomical knowledge. Two automated pathway mapping methodologies have been developed to address this challenge: volumetric CN segmentation and fiber clustering. Volumetric CN analysis has emerged as an effective strategy for CN tract segmentation across different MRI modalities, including T1-weighted (T1w) images, T2-weighted (T2w) images, fractional anisotropy (FA) images and fiber orientation distribution (FOD) peak images, which directly classify voxels based on associated fiber bundles, thereby circumventing conventional streamline analysis~\cite{diakite2025dual, zeng2025rgvpseg}. Alternatively, single-pair CN atlases were created by fiber clustering to automatically map the pathways, such as CN II atlas~\cite{zeng2023automated}, CN III atlas~\cite{huang2022automatic}, CN V atlas~\cite{2020Creation}, CN VII/VIII atlas~\cite{zeng2021automated}, which obtained fiber clustering maps by analysing the spatial distribution and distance features of different fiber bundles. Furthermore, Li et al.~\cite{li2024tractography} advanced this field by proposing a microstructure-informed supervised contrastive learning framework for CN II pathway identification, integrating streamline labels with tissue microstructure data to determine positive and negative pairs. However, in the context of skull base surgery, single-pair CN atlas or volumetric analysis struggle to achieve a complete description of multiple pairs of involved nerves around the tumor. Furthermore, the automated identification workflow for single-pair CN atlas involves computationally intensive steps such as multi-subject fiber registration and spectral clustering, which compromise clinical applicability by failing to meet intraoperative requirements for simultaneous high efficiency and precision.

By addressing these limitations, this study aims to develop a comprehensive diffusion tractography atlas encompassing multiple CN pairs to enable automated pathway mapping in the human brain.
The complex skull base environment significantly complicates fiber tractography, as uniform parameters across different CN pairs frequently yield suboptimal results. For instance, optimal parameters for CN II pathway reconstruction using the two-tensor Unscented Kalman Filter (UKF) approach~\cite{zeng2023automated} prove ineffective for CN V tractography~\cite{2020Anatomical}. Therefore, we seek to exploit a multi-parametric diffusion tractography atlas to cater to each pair of CNs to obtain the most anatomically correct tracking results. Additionally, anatomical variations among CN pairs create divergent inter-streamline distances, challenging conventional clustering methods in eliminating false-positive fibers. Instead of disposable clustering, we explore new multi-stage fiber clustering strategy to reduce the generation of streamlines that do not correspond to anatomical locations. By integrating these advancements, we aim to construct a multi-parametric multi-stage diffusion tractography atlas for mapping the pathways of five pairs of CNs.
 
 In this study, we propose a comprehensive dMRI tractography atlas of cranial nerves (CNs) capable of automatically mapping 8 fiber bundles associated with 5 pairs of CNs: CN II, III, V, and VII/VIII. First, we employ multi-fiber UKF tractography to reconstruct 3D CN trajectories, with parameter optimization tailored to each CN pair's distinct anatomy. The CN atlas was generated via a multi-stage fiber clustering approach using dMRI data from 50 healthy subjects in the Human Connectome Project (HCP). Finally, we validated the atlas's generalizability by automating pathway mapping across diverse datasets: HCP, multi-shell dMRI (MDM), pituitary adenoma (PA) patients, and a craniopharyngioma (CP) patient. Qualitative and quantitative experimental results demonstrate that the proposed method has ideal colocalisation with expert manual identification. Both the implementation code and the atlas data are publicly available. We hope that this resource will enhance the standardisation of CN neuroimaging analysis and welcome efforts to improve the accuracy of the atlas and extend it to other CNs. The main contributions are summarised as follows: 
\begin{itemize}
	\item We create a comprehensive dMRI tractography atlas for automatically mapping the pathways of 5 pairs of CNs, including CN II, CN III, CN V and CN VII/VIII.
	\item We introduce two technical novelties in our proposed atlas. First, a multi-parametric diffusion tractography strategy is designed to cater to each pair of CNs to obtain the most anatomically correct tracking results. Second, a novel multi-stage fiber clustering is proposed to reduce the generation of streamlines that do not correspond to anatomical locations.
	\item Extensive experimental results show that the proposed method has ideal colocalisation with expert manual identification on the HCP dataset, the MDM dataset, PA patients and a CP patient.
\end{itemize}

\section{Related Work}
\label{sec:rw}
\subsection{CN Pathways Mapping}\label{sec:CA}
Early approaches~\cite{sultana2017mri} identified the CN pathways from T1w and T2w images, relying on neurosurgeons to manually label approximate regions or apply specific deformation models. With the advancements in diffusion MRI, mapping the pathways of CNs has been performed by resolving fiber orientation in each voxel, constructing appropriate streamline reconstruction methods and using specific strategies to obtain anatomically accurate clusters of streamlines. For example, Yoshino et al. \cite{yoshino2016visualization} employed high-definition fiber tractography (HDFT) to map the cisternal segments of most CNs in healthy subjects. Jacquesson et al. \cite{jacquesson2019probabilistic} demonstrated how probabilistic CN tractography could inform surgical planning for diverse skull base tumors. Similarly, Zolal et al.~\cite{zolal2016comparison}, He et al.~\cite{he2021comparison}, and Xie et al.~\cite{2020Anatomical} systematically compared fiber tractography methods, evaluating various region-of-interest (ROI) filtering strategies to extract anatomically plausible CN fiber bundles. owever, current CN identification in dMRI tractography depends on manual ROI selection, requiring surgeons to interactively place ROIs—a time-intensive and clinically inefficient process. Thus, automated CN pathway mapping is essential for enhancing both efficiency and precision in clinical applications.

\subsection{Fiber Clustering}
Fiber clustering methods are widely employed in white matter mapping and visualization, where tractography streamlines are grouped by geometric trajectories to characterize structural connectivity based on white matter anatomy~\cite{O'Donnell20071562,ODONNELL2013283}. To overcome manual ROI placement, fiber clustering methods enable automated identification of CN pairs based on their distinct anatomical structures. For example, Zhang et al.~\cite{2020Creation} developed a CN V tractography atlas for identifying three components: the cisternal segment, mesencephalic trigeminal tract, and spinal trigeminal tract. Huang et al.~\cite{huang2022automatic} proposed a data-driven fiber clustering strategy to map three kinds of positional relationships with the red nuclei and two kinds of positional relationships with medial longitudinal fasciculus. Similarly, Zeng et al.~\cite{zeng2021automated} constructed the identification framework for CN VII/VIII tracts that enter the cisternal portion through the cerebellopontine angle, on the basis of which the mapping of the two decussating and two nondecussating pathways of the CN II was also implemented~\cite{zeng2023automated}. However, existing fiber clustering algorithms cannot meet the needs of streamline screening for multiple pairs of CNs with different anatomy. Therefore, how to improve the fiber clustering strategy to create a comprehensive dMRI tractography atlas that simultaneously maps multiple pairs of CN pathways is the key to obtain the spatial relationships of multiple pairs of involved nerves around the tumor preoperatively for skull base surgery.
\begin{figure*}[]
	\centering
	\includegraphics[width=0.92\textwidth]{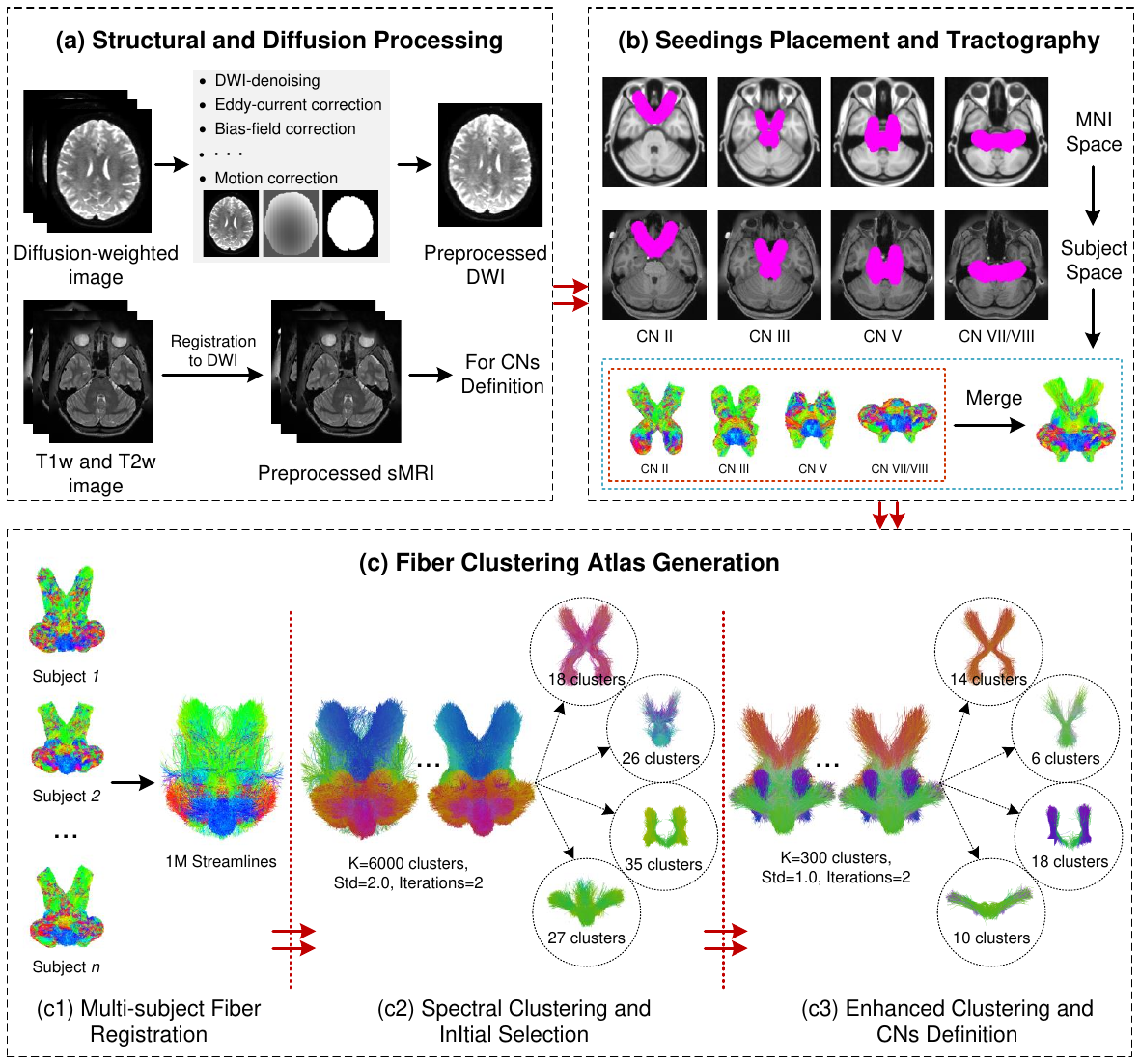}
	\caption{Overview of CN atlas generation pipelines. (a) Structural and diffusion MRI were passed through corresponding preprocessing steps for anatomical definition and tracking. (b) Fiber tractography for the reconstruction of CN pathways from the seeding images. (c) Multi-stage fiber clustering atlas generation using tractography data from HCP dataset.}
	\label{fig:2}
\end{figure*} 
\section{METHOD}
\label{sec:METHODOLOGY}
\subsection{Datasets and Preprocessing}
A total of 110 cases of HCP dataset~\cite{sotiropoulos2013advances,van2013wu} were used in this paper, 50 cases to generate CN atlas, and the remaining 60 to validate the proposed method. Meanwhile, we apply our method to 20 cases of MDM~\cite{tong2020multicenter} dataset, two PA patients~\cite{zeng2023automated}, and a CP patient. The detailed parameters of the dMRI data are as follows:

\textit{HCP dataset}: The HCP provides high-quality dMRI, T1w images, and T2w images, which are approved by the local Institutional Review Board of Washington University. The dMRI acquisition parameters in HCP are as follows: 18 base images with b-values=0 $s/mm^2$ and 270 gradient directions. b=1000, 2000, and 3000 $s/mm^2$, TR=5520 $ms$, TE=89.5 $ms$, matrix size=145$\times$174$\times$145 $mm^3$, resolution=1.25$\times$1.25$\times$1.25 $mm^3$ voxels. The T1w images acquisition parameters are as follows: TR=2400 $ms$,, TE=2.14 $ms$,, TI=700,2500 $ms$,, matrix size =145$\times$174$\times$145 $mm^3$, voxel size=1.25$\times$1.25$\times$1.25 $mm^3$. 

\textit{MDM dataset}: The MRI data of MDM dataset was collected from three traveling subjects with identical acquisition settings in 10 imaging centers. The dMRI acquisition parameters are as follows: 6 base images with b-values=0 $s/mm^2$ and 90 gradient directions with other three b-values of 1000, 2000, and 3000 $s/mm^2$, TR=5400 $ms$, TE=71 $ms$, FOV=220$\times$220 $mm^2$, slice number=93, voxel size=1.5$\times$1.5$\times$1.5 $mm^3$. The T1w images acquisition parameters are as follows: TR=5000 $ms$, TE=29 $ms$, TI=700,2500 $ms$, FOV=211$\times$256$\times$256 $mm^3$, voxel size=1.2$\times$1$\times$1 $mm^3$

\textit{PA patients}: The MRI data of the PA patients are acquired at Xuanwu Hospital Capital Medical University by using the Siemens Skyra 3T scanner. The dMRI acquisition parameters in PA patient data are: 60 gradient directions and b-values=1000 $s/mm^2$, TR=8900 $ms$, TE=95 $ms$, voxel size=1.8$\times$1.8$\times$2.8 $mm^3$. The T1w images acquisition parameters are as follows: TR=2400 $ms$, TE=2.27 $ms$, FOV=250$\times$250 $mm^2$, voxel size=1.0$\times$1.0$\times$1.0 $mm^3$.

\textit{The CP patient}: The MRI data of the CP patient are acquired at Xuanwu Hospital Capital Medical University by using the Siemens Skyra 3T scanner. The dMRI acquisition parameters in CP patient data are: 60 gradient directions and b-values=1000 $s/mm^2$, TR=7400 $ms$, TE=102 $ms$, Voxel size=1.8$\times$1.8$\times$2.8 $mm^3$. The T1w images acquisition parameters are as follows: TR=2400 $ms$, TE=2.41 $ms$, FOV=256$\times$256 $mm^2$, Slice number=176, Voxel size=1.0$\times$1.0$\times$1.0 $mm^3$.

The HCP dataset, MDM dataset and patient data were pre-processed with the standard DWI pre-processing pipeline, including motion correction, eddy current correction and EPI distortion correction. For HCP data, we used dMRI data that were already pre-processed with the HCP minimum processing pipeline (\textit{https://github.com/pnlbwh/pnlpipe/tree/v2.2.0}). For the MDM dataset, two PA patients and one CP patient, we performed standard pre-processing, including denoising, motion correction, eddy current correction, EPI distortion correction and co-registration of the FMRIB Software Library (FSL) (\textit{https://fsl.fmrib.ox.ac.uk/fsl})\cite{jenkinson2012fsl}. In these dMRI datasets, we used the \textit{dwiextract} command of the Mrtrix3~\cite{tournier2012mrtrix} (\textit{https://www.mrtrix.org}) to extract single-shell dMRI data to perform fiber tractography to save computational time and memory.

\subsection{Seeding Images Placement}
To facilitate the creation of subject-specific tractography masks, we designed a semi-automated process based on linear registration. As shown in Fig.~\ref{fig:2}(b), the main steps are as follows: Firstly, we manually drew a mask that covers all CNs on the T1 template in MNI space~\cite{grabner2006symmetric}. Secondly, we registered this mask region to each individual subject on the basis of linear registration between the subject’s T1w image and the T1 template of the MNI space by using \textit{FIRST} command from the FSL~\cite{jenkinson2012fsl}, which is a model-based registration for automatic segmentation of a number of subcortical structures. Finally, we transferred the mask to the dMRI images in the subject’s individual space. We visually inspected the registration of all subject masks. Considering that anatomical a priori knowledge from tractography influences the quality of CN reconstruction, we set seed points covering each of the five pairs of CNs in MNI space. The specific anatomical details are as follows: i) CN II begins at the retina, passes through the bilateral optic nerves, overlies the optic chiasm, converges on the contralateral optic tract and finally extends to the lateral geniculate nucleus of the thalamus; ii) CN III starts from interpeduncular fossa and extends outward to the cavernous sinus; iii) CN V exits the brainstem from the junction of the cerebral bridge and the cerebral bridge arm; and iv) CN VII/VIII starts at cerebellar peduncles and ends at the internal auditory canal.
\subsection{Multi-parametric Multi-tensor CNs Tractography}
These studies~\cite{he2021comparison,2020Anatomical,epprecht2024facial} demonstrated the effectiveness of the UKF method compared with the FOD method for CN tractography in reducing false-positive fiber production. In this paper, we select the two-tensor UKF method (\textit{https://github.com/pnlbwh/ukftractography}) to map the three-dimensional trajectory of CN pathways. For each pair of CNs with unique anatomical structures, we design a multi-parametric multi-tensor tractography method, which sets the different tractography parameters for fiber tracking derived from the optimal parameters provided in~\cite{he2021comparison,huang2022automatic,2020Anatomical,zeng2021automated}. As shown in Fig.~\ref{fig:2}(b), after the seed images in MNI space are registered to the individual space, we use them as seed points and restriction regions for individual tracking. The details of the CN tractography parameters are as follows: i) CN II: \textit{seedingFA}=0.02, \textit{stoppingFA}=0.01, \textit{Qm} (rate of change of tensor direction)=0.001, and \textit{Ql} (rate of change of eigenvalues)=50; ii) CN III: \textit{seedingFA}=0.01, \textit{stoppingFA}=0.01, \textit{Qm}=0.001 and \textit{Ql}=150; iii) CN V: \textit{seedingFA}=0.06, \textit{stoppingFA}=0.05, \textit{Qm}=0.001, and \textit{Ql}=300; iv) CN VII/VIII: \textit{seedingFA}=0.02, \textit{stoppingFA}=0.05, \textit{Qm}=0.001 and \textit{Ql}=50. Six seeds per voxel were used for seeding the CN tractography, which resulted in about 50,000 fibers in the tractography of each pair of CNs for each subject. Finally, the fiber streamlines of the five pairs of CNs were merged, resulting in approximately 200,000 fibers in the tractography for each subject.
\subsection{Multi-stage Fiber Clustering Atlas Generation}
As shown in Fig.~\ref{fig:2}(c), a comprehensive diffusion tractography atlas was generated using data-driven fiber clustering pipeline, as implemented in the \textit{whitematteranalysis} software (\textit{https://github.com/SlicerDMRI/whitematteranalysis}). The pipeline consists of three key steps: multi-subject fiber registration was used to register the tractography of all subjects to a common space, spectral clustering of fiber streamlines was performed to subdivide the registered tractography data into multiple fiber clusters simultaneously and anatomical tract definition was applied to identify each pair of CN clusters.
\subsubsection{Multi-subject Fiber Registration}
To register the tractography of all subjects to a common space (Fig.~\ref{fig:2}(c)), we perform an unbiased entropy-based groupwise tractography registration of CN tractography from each subject to create a high-dimensional fiber space~\cite{o2012unbiased}. Specifically, we analyse streamlines by randomly sampling 20,000 fibers from each individual. A total of 1 million fibers were registered into a common space with a minimum fiber length of 20 mm, followed by affine and coarse-to-fine b-spline registration with multiscale sigma values from 20 down to 2 mm and a final b-spline grid size of 8$\times$8$\times$8. After multi-subject fiber tractography registration, a high-dimensional CN tractography is obtained for multi-stage fiber clustering.
\subsubsection{Multi-stage Fiber Clustering}
Instead of normal spectral clustering, we explore a new strategy of multi-stage fiber clustering to reduce the generation of streamlines that do not correspond to anatomical positions, which consists of initial spectral clustering and enhanced fiber clustering. Firstly, we randomly sample 20,000 fibers from the registered tractography of each subject, for a total of 1 million fibers. Notably, many tractography fibers are highly similar to their neighbouring fibers on the basis of anatomical principles. Thus, we perform a random sampling of 20,000 fibers from each subject rather than analysing all fibers across subjects and ensure that the extracted number of fibers is sufficient to represent the anatomical structure of the cranial nerve with limited computational effort~\cite{2020Creation}. As shown in Fig.~\ref{fig:2}(c2), for initial spectral clustering, CN tractography is divided into K = 6,000 clusters by spectral clustering to create a fiber clustering atlas. In this paper, we perform a coarse classification by dividing the results from the result of multi-subject fiber registration into 6,000 clusters in the initial spectral clustering. We generate multiple atlases of different scales by setting different K values (K = 3000, 4000, 5000 and 6000). The results show that when the atlas adopts a coarse classification scale (K\textless6000), it can distinguish fiber clusters of the cranial nerve but includes many false-positive fibers. Considering the issues of subsequent anatomical labelling and computational cost, we did not select a higher number of classifications. On the basis of the characteristics of anatomical and distance similarity consistency between fibers~\cite{O'Donnell20071562}, we classify the 1 million fibers in the high-dimensional fiber space into 6,000 clusters by using a measure of pairwise distances. Then, we perform two iterations (standard deviations: 2.0) on the clustering results to remove false-positive fibers in each cluster. Given the chosen CN fiber clustering atlas, each fiber cluster was initially selected to indicate whether it belongs to the CNs or not, as obtained by the initial screening in which ROIs on T1w images and T2w images were placed in the atlas space. In this way, 106 CNs clusters in the atlas (Fig.~\ref{fig:2}(c2)) were obtained, each representing a specific anatomical branch of CNs, including CN II (18 clusters), CN III (26 clusters), CN V (35 clusters) and CN VII/VIII (27 clusters). 
After spectral clustering and initial selection, the resulting clusters of fibers that do not fit the anatomical definition of CNs may still exist. Therefore, enhanced clustering is used to automatically select the correct fibers again from the above clusters results. Specifically, we merge 106 fiber clusters in the initial atlas into a high-dimensional fiber tractography that comprises approximately 30,000 streamlines. From this set, we randomly select 20,000 streamlines to maintain consistency with the number selected in the first stage while ensuring the fibers are sufficiently representative of the target anatomy. During the process of enhanced fiber clustering, we test cases such as K = 100, 200 and 300. The results show that the fiber clusters of the cranial nerve can be sorted out when the atlas chooses a coarse classification scale (K = 100), but many false-positive fibers are included. When we choose a small classification scale (K = 200), fiber clusters that belong to the cranial nerve hardly present false-positive fibers and differentiate the anatomical differences between different CNs. In addition, we set strict standard deviations (standard deviations: 1.0) from the cluster’s mean fiber affinity to remove certain outlier fibers, where certain fibers are outliers if they are far away from other fibers in the cluster in which they are located. Next, all 200 candidate clusters in enhanced clustering atlas were defined one by one in terms of whether they belonged to CNs or not by expert rater manual annotation. Specifically, the T1w images and T2w images of the 50 subjects from HCP for which the atlas was created were registered into the atlas space to obtain the population-mean T1w image and T2w image, which will be used as a reference for the experts. Another expert rater viewed the curated CN clusters to confirm their anatomical correctness. Overall, 74 clusters in the enhanced atlas (Fig.~\ref{fig:2}(c3)) were selected, which represent five pairs of CNs, including CN II (14 clusters), CN III (6 clusters), CN V (18 clusters) and CN VII/VIII (10 clusters). 
\subsection{Automated CN Identification of New Subjects}\label{sec:New Subjects}
We apply the created multi-stage fiber atlas to automatically map the pathways of CNs of a new subject. Firstly, we use the well-established diffusion and structural processing pipeline to handle a new subject (Fig.~\ref{fig:2}(a)). We then register the seeding and mask in the MNI space to the individual subject, perform fiber tractography for each pair of CNs and merge to obtain a high-dimensional fiber streamline that is representative of all pairs of CNs. Secondly, the obtained fiber streamline is registered into the CN atlas space by affine transformation and non-rigid registration. Thirdly, each registered streamline from subject-specific CN tractography is assigned to its closest cluster of multi-stage fiber atlas. Outlier fibers are removed using the same parameters as those used to create the atlas. Finally, automated CN identification of new subjects is performed by finding the subject-specific clusters that correspond to the defined atlas clusters.

\section{Experimental Evaluation}
\label{sec:EXPERIMENTS}
In this work, we test the proposed CN atlas on different acquirement sites, including the HCP dataset, MDM dataset, PA patients and a CP patient. Quantitative and visual experiments are performed using dMRI data and compared with manual CN selection, which is filtered by setting ROIs and ROAs. For quantitative comparisons, we choose validation metrics for the commonly used atlas to compare the proposed automated CN identification method with manual CN selection, including CN identification rate and CN spatial overlap~\cite{zeng2023automated}.

\begin{table*}[]
	\centering
	\caption{CN identification rate of each pair of CNs using the proposed CN atlas and the manual selection method.}
	\label{tab:1}
	\resizebox{0.8\textwidth}{!}{%
		\begin{tabular}{c||cccccccc}
			\hline\hline
			\multirow{2}{*}{Dataset}  & \multicolumn{2}{c}{CN II-D}& \multicolumn{2}{c}{CN II-N }& \multicolumn{2}{c}{CN III-L}& \multicolumn{2}{c}{CN III-R} \\ 
			& Auto.           & Manu.  & Auto.           & Manu.  & Auto.           & Manu.& Auto.           & Manu.   
			\\\hline
			
			HCP subjects (n=60) Successful subjects
			&50/50	&50/50	&49/49	&49/49   &	54/54	&54/54	&56/56&	56/56
			\\
			
			Unsuccessful subjects &8/10&	0/10&	11/11&	0/11&	5/6&	0/6&	4/4&	0/4
			\\\hline
			MDM subjects (n=20) Successful subjects &18/18&	18/18&	10/10&	10/10&	18/19	&19/19&	18/19&	19/19
			\\
			
			Unsuccessful subjects &1/2&	0/2&	3/10&	0/10&	0/1&	0/1	&0/1&	0/1	
			\\\hline
			PA patients (n=2) &2/2&	2/2	&2/2&	2/2&	1/2&	1/2&	2/2&	2/2
			\\\hline
			CP patient (n=1) &1/1&	1/1	&1/1&	1/1	&	1/1	&	1/1	&	1/1	&	1/1	
			\\\hline\hline
			\multirow{2}{*}{Dataset}  & \multicolumn{2}{c}{CN V-L}& \multicolumn{2}{c}{CN V-R }& \multicolumn{2}{c}{CN VII/VIII-L}& \multicolumn{2}{c}{CN VII/VIII-R} \\ 
			& Auto.           & Manu. & Auto.           & Manu.& Auto.           & Manu.& Auto.           & Manu.   
			\\\hline
			HCP subjects (n=60) Successful subjects
			&51/51	&51/51&	57/57&	57/57&	54/54&	54/54	&54/54	&54/54
			\\
			
			Unsuccessful subjects &5/9&	0/9	&3/3&	0/3&	4/6&	0/6&	6/6&	0/6
			\\\hline
			MDM subjects (n=20) Successful subjects &13/13	&13/13&	10/11&	11/11&	14/17&	17/17&	17/20	&20/20
			\\
			
			Unsuccessful subjects &2/7&	0/7	&5/9&	0/9&	2/3&	0/3&	0/0&	0/0	
			\\\hline
			PA patients (n=2) &2/2&2/2	&2/2&2/2&2/2&2/2&2/2&2/2
			\\\hline
			CP patient (n=1) &1/1	&	1/1	&1/1	&	1/1	&1/1	&1/1	&0/1	&0/1	
			
			\\ \hline\hline
			
		\end{tabular}
	}
\end{table*}
\begin{table*}[]
	\centering
	\caption{Spatial overlap (wDice) between the proposed CN atlas and the manual selection method.}
	\label{tab:2}
	\resizebox{0.75\textwidth}{!}{%
		\begin{tabular}{c||ccccc}
			\hline\hline
			
			Datasets &CN II	&CN III	&CN V	&CN VII/VIII &	5 pairs of CNs\\\hline
			HCP subjects &0.7445±0.0915	&0.7296±0.1107	&0.7849±0.1467	&0.7203±0.0975	&0.7448±0.0546\\
			MDM subjects &0.7806±0.1551	&0.7956±0.0891	&0.8787±0.1501	&0.6760±0.1363&	0.7827±0.0477
			
			\\ \hline\hline
			
		\end{tabular}
	}
\end{table*}
\begin{figure*}[]
	\centering
	\includegraphics[width=0.8\textwidth]{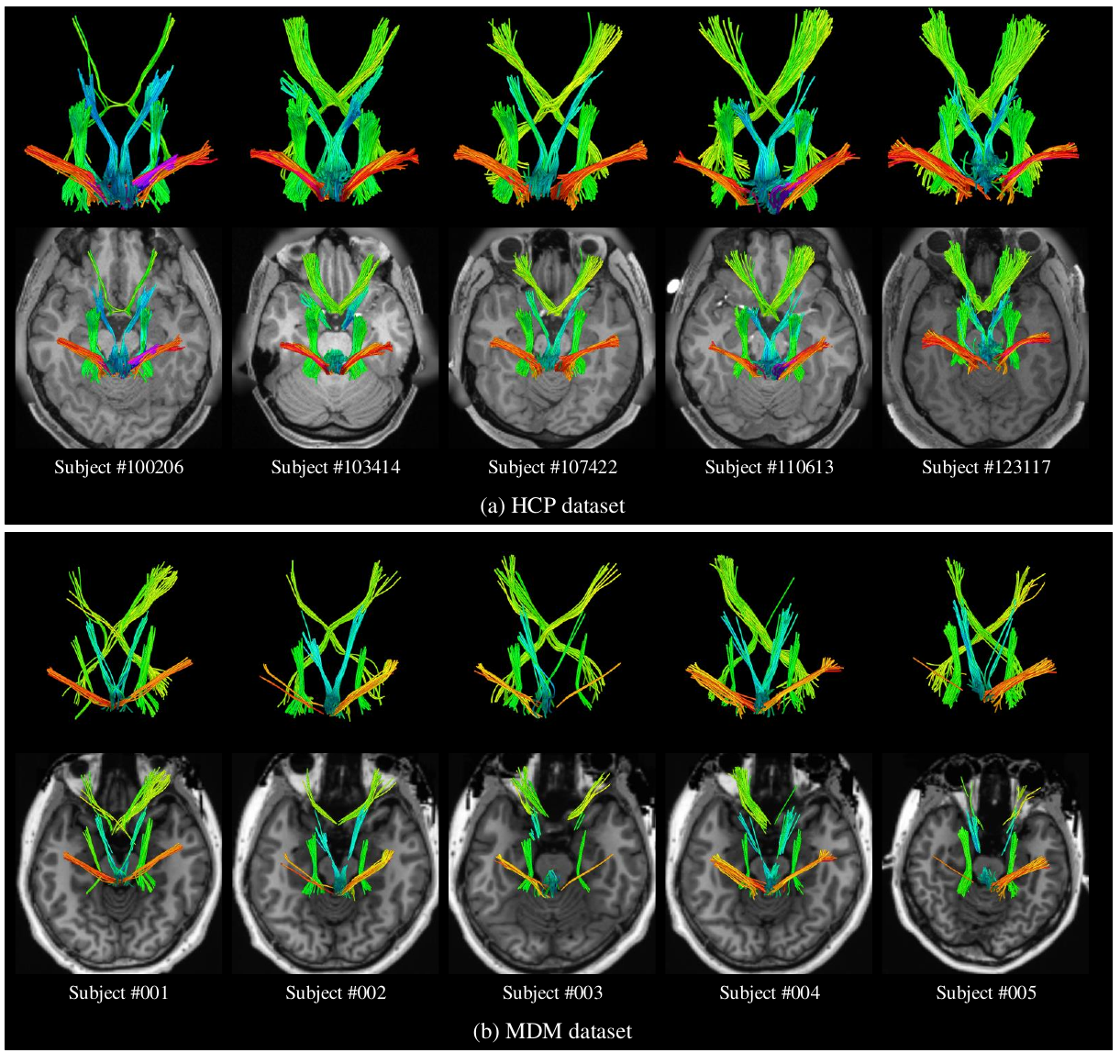}
	\caption{Examples of the CN pathway reconstruction results using the proposed CN atlas on HCP dataset and MDM dataset. The tractogram is color-coded in average directional color using the 'Color Fibers By Mean Orientation' buttons of the \textit{TractographyDisplay} Module of the 3D Slicer.}
	\label{fig:3}
\end{figure*}
\subsection{Identification of Ground Truth CN using Manual Selection}
For the testing subjects, we perform manual ROI-based CN identification, which is filtered by setting ROIs and ROAs in Supplementary Materials of Ref~\cite{xie2023cntseg}, and use them as ground truth for evaluating automated CN identification. For each pair of CNs, we set different ROIs and ROAs to select and filter as follows: i) CN II: ROI-1 was drawn at the posterior part of the bilateral optic tracts of the CN II from the coronal view, ROI-2 in the optic chiasm was drawn from the coronal view and ROI-3 was placed at the anterior part of the bilateral optic nerves on the diffusion tensor map from the coronal view. ii) CN III: ROI-1 was drawn at the cisternal segment near the cavernous sinus, ROI-2 was drawn at the mid of the CN III cisternal segment, ROI-3 was placed at the CN III cisternal segment near interpeduncular fossa and ROI-4 was placed at the mesencephalon. One exclusion ROI in the midline was drawn for every subject. iii) CN V: ROI-1 was drawn at the cisternal portion of CN V, and ROI-2 in Meckel’s cave (MC) was drawn on the mean b=0 image. One exclusion ROI in the midline was drawn for every subject. iv) CN VII/VIII: ROI-1 was placed at the cerebellopontine angle of the CN VII/VIII, and ROI-2 was placed at the internal auditory canal of the CN VII/VIII. ROA-1 was used for exclusion of fibers in the whole brain except the brainstem.
\subsection{Validation Metrics}
We compute the mean CN identification rate and spatial overlap across all subjects in each dataset. For the CN identification rate, we perform the evaluation for each pair of CNs, including the decussating (CN II-D) and nondecussating (CN II-N) tracts of the CN II, the left (CN III-L) and right (CN III-R) tracts of the CN III, the left (CN V-L) and right (CN V-R) tracts of the CN V, and the left (CN VII/VIII-L) and right (CN VII/VIII-R) tracts of the CN VII/VIII. We report the mean identification rate of the 60 HCP testing subjects and the 20 MDM testing subjects, and for subjects where manual ROI-based CN identification was not successful, we compare this with the proposed automated method. For CN spatial overlap, we compute it to assess if the CNs that were identified using the proposed automated method spatially overlapped with the manually identified CNs. We choose wDice~\cite{cousineau2017test,zhang2022quantitative,zhang2019test} as the metric for evaluating spatial overlap, which was designed specifically for measuring volumetric overlap of fiber tracts. For all testing datasets, we report the mean and the standard deviation of wDice with successful manual CN selection.
\section{RESULTS}
\label{sec:RESULTS}
\subsection{CN Identification Rate}
Table~\ref{tab:1} presents the CN identification rate of different subdivisions of each CN by using the proposed CN atlas and the manual selection method. For a clearer comparison, we give the corresponding identification performance of the proposed CN atlas on the basis of whether the manual method was able to identify different branches of the CNs. Among the subjects whose CNs were successfully identified by the manual selection method on the HCP subjects and MDM subjects, almost all CNs were identified by the proposed CN atlas. Even in subjects whose CNs were unsuccessfully identified by the manual selection method, most CNs were identified by the proposed CN atlas. For instance, in subjects from HCP whose CN III-L (0/6) and CN III-R (0/4) tracts of CN III were unsuccessfully identified by the manual ROI-based method, some CN III-L (5/6) and CN III-R (4/4) tracts were identified by the proposed CN atlas.
\subsection{CN Spatial Overlap}
Table~\ref{tab:2} presents the mean and the standard deviation of the wDice scores across HCP subjects and MDM subjects with successful manual CN identification (subjects in which each pair of CNs was successfully identified). As can be seen in Table~\ref{tab:2}, the mean wDice of automatic identification of CN II, CN III, CN V and CN VII/VIII from the proposed CN atlas reaches 0.7445, 0.7296, 0.7849 and 0.7203, respectively. The overall mean wDice of all CNs can reach 0.7448, which is higher than the standard threshold ($\geq$0.72) for evaluating fiber spatial overlap proposed by~\cite{cousineau2017test}. For MDM subjects, the CN II, CN III, CN V, CN VII/VIII and all CNs obtained from the proposed CN atlas reach 0.7806, 0.7956, 0.8787, 0.6760 and 0.7827, respectively. These results show a high CN spatial overlap between the proposed CN atlas and the manual selection method.
\begin{figure}[t]
	\centering
	\includegraphics[width=0.46\textwidth]{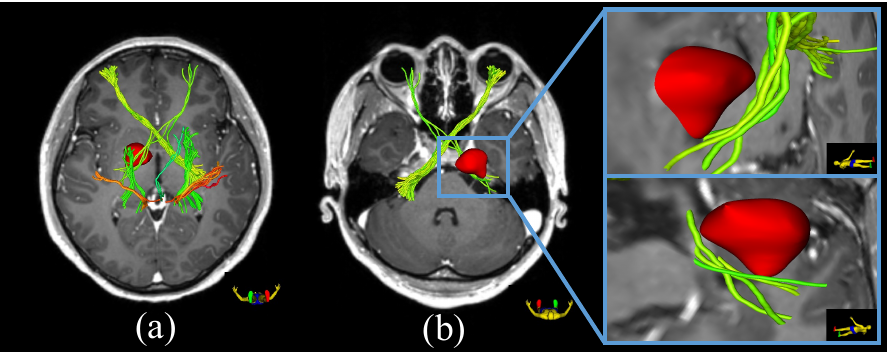}
	\caption{Reconstruction results of the proposed CN atlas in a 28-year-old woman PA patient data. (a) Mapping results of 5 pairs of CNs pathways overlaid on the transverse plane of the T1w image. (b) Mapping results of the CN II pathway surrounding the tumor,  and the tumor is red color. The upper right and lower right panels are magnified views of different perspectives of the relationship between the tumor and the nerve pathway in selected areas. }
	\label{fig:4}
\end{figure}
\begin{figure}[]
	\centering
	\includegraphics[width=0.46\textwidth]{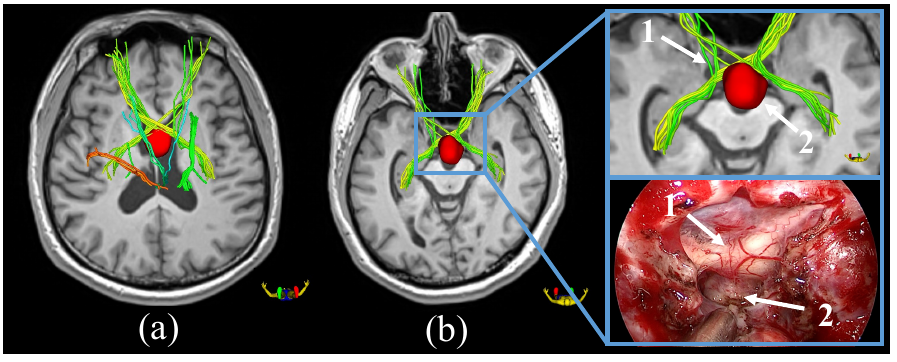}
	\caption{Reconstruction results of the proposed CN atlas in a 52-year-old man CP patient data with intracranial space-occupying lesions. (a) Mapping results of 5 pairs of CNs pathways overlaid on the transverse plane of the T1w image. (b) Mapping results of the CN II pathway surrounding the lesions. The upper right panel displays a magnified image of the selected area, and the lower right panel shows the corresponding intraoperative view. 1 and 2 are the optic chiasm of the CN II and lesions, respectively.}
	\label{fig:5}
\end{figure}
\subsection{CN Visualization}
To demonstrate the identification performance of the proposed CN atlas, we select different CN tracts from different subjects of the HCP dataset and the MDM dataset for qualitative comparison. Fig.~\ref{fig:3}(a) gives the CN pathway reconstruction results of subjects \#100206, subjects \#103414, subjects \#107422, subjects \#110613 and subjects \#123117 using the proposed CN atlas on the HCP dataset, respectively. Fig.~\ref{fig:3}(b) shows the CN pathway reconstruction results of subjects \#001, subjects \#002, subjects \#003, subjects \#004 and subjects \#005 using the proposed CN atlas on the MDM dataset, respectively. The top and bottom rows of Figs.~\ref{fig:3}(a) and ~\ref{fig:3}(b) provide the 3D views of the identified CNs and displays of the 3D view superimposed on the slice, respectively. The CNs identified by the proposed CN atlas conform to the anatomical shape and fit the anatomical location on the slice.
\subsection{Performance on two PA Patients}
In this experiment, we select two PA patients and use the proposed CN atlas to automatically identify the CN II surrounding the tumor. Regarding identification rate, Table~\ref{tab:1} indicates that the proposed CN atlas successfully identified all CN II-D (2/2) and CN II-N (2/2) tracts of CN II subdivisions. For CN II visualization, Fig.~\ref{fig:4} shows the different slices of the reconstruction results of the CN II pathway surrounding the tumor. From these results, we can find that the proposed CN atlas successfully identified different CN II subdivisions that surround the tumor.

\subsection{Performance on CP Patient}
In this experiment, we select a 52-year-old male CP patient to test the performance of the proposed CN atlas and to perform further validation with intraoperative pictures. For the identification rate, Table~\ref{tab:1} shows that the proposed CN atlas successfully identified all CN tracts except for CN VII/VIII-R. For CN II surrounding the tumor, the proposed method demonstrated superior pathway mapping. Fig.~\ref{fig:5} shows the reconstructed CN pathways around the tumor, along with the corresponding intraoperative images. The CN pathway mappings generated by the proposed atlas were consistent with intraoperatively observed locations.
\begin{figure}[b]
	\small
	\centerline{
		\begin{tabular}{@{}c@{}c@{}c@{}c@{}c@{}c@{}}
			
			\includegraphics[width=0.2\textwidth]{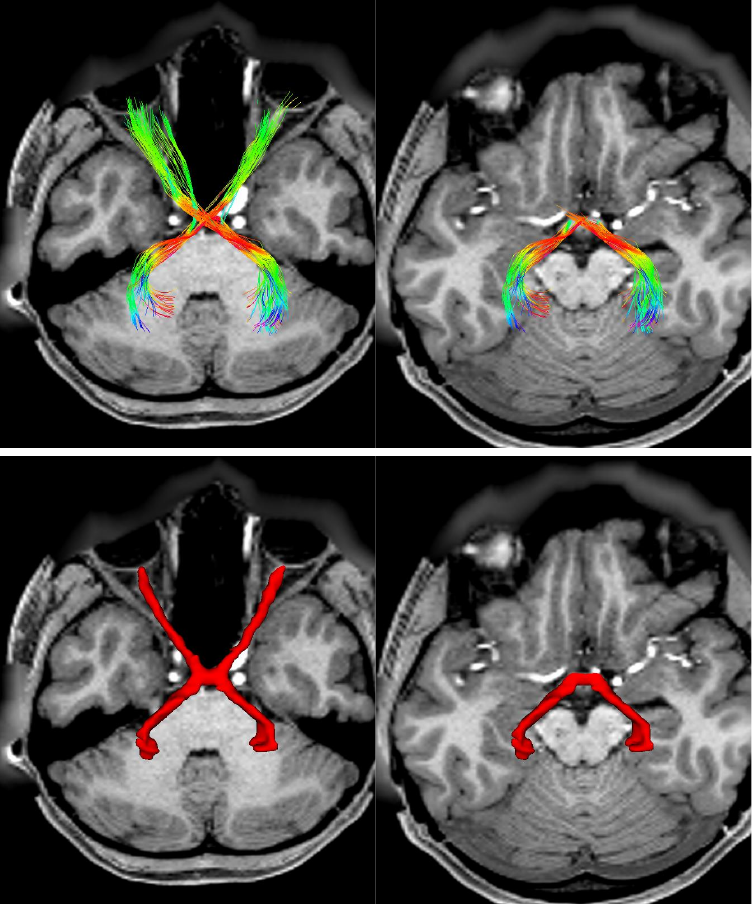}
			& \includegraphics[width=0.2\textwidth]{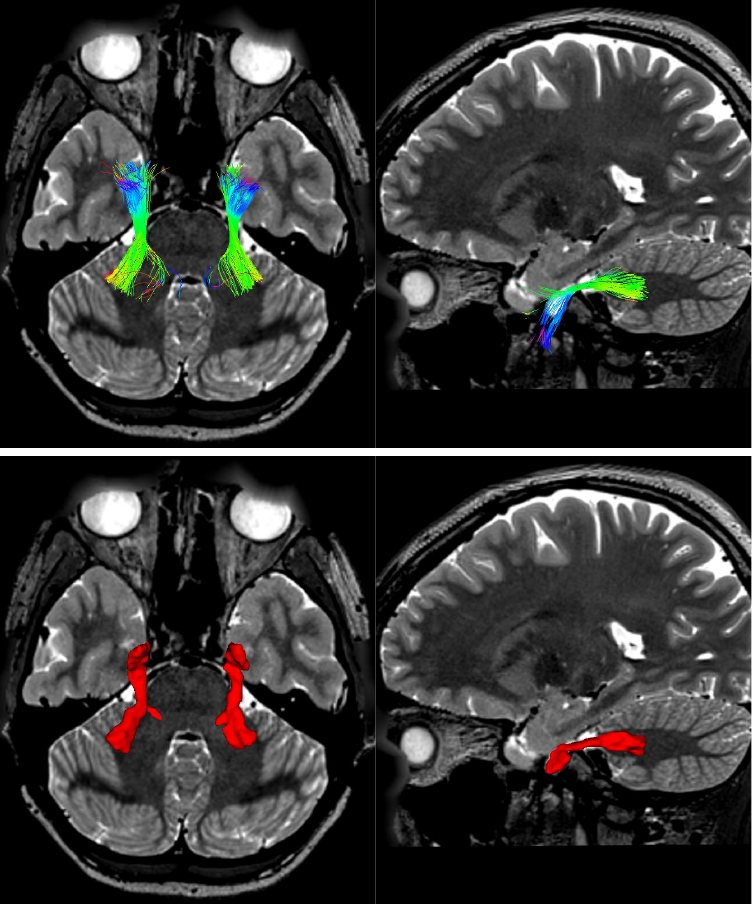}
			\\
			CN II & CN V    
	\end{tabular}}
	\caption{CN reconstruction results of CNTSeg and our proposed CN atlas. The top image shows our proposed CN atlas, where the fiber streamlines are color-coded based on their local orientation. The bottom image displays 3D rendering of the volumetric segmentation model CNTSeg.}\label{tract_seg}
\end{figure}
\subsection{Vs. Volumetric Cranial Nerve Segmentation}
Volumetric cranial nerve segmentation, e.g. CNTSeg~\cite{xie2023cntseg}, which is an initial exploration of label assignment to each voxel by using multimodal fusion, exhibits promising segmentation performance. In this paper, our proposed CN atlas involves analysing the generated streamlines and delineating the corresponding fiber bundles on the basis of anatomical definitions. To facilitate a visual comparison between volumetric segmentation and CN tractography atlas, we present the streamline generated by the CN atlas and the region segmented by CNTSeg. Fig.~\ref{tract_seg} displays the CN II and CN V reconstruction results of the CNTSeg and CN tractography atlas on the HCP subject. From Fig.~\ref{tract_seg}, we can see an overall consistency in the CNs generated by the CNTSeg and CN atlas, with CN atlas generating more elongated streamlines in the brainstem and the CNTSeg generating results with fewer false positives in the extracranial regions. Both methods achieve consistent overall CN identification, yet they have critical differences: The atlas method produces anatomically plausible streamlines by enforcing spatial continuity (e.g. CN II optic tract continuity), whereas CNTSeg directly localises neural pathways on a voxel-by-voxel basis. These two approaches are complementary, and future research could integrate both to achieve a fast, accurate and robust approach to CN analysis.

\section{Discussion}\label{sec:Discussion}
This work presents a multi-parametric, multi-stage diffusion tractography atlas for automated mapping of CN pathways in the human brain. Our method demonstrates CN identification performance comparable to expert manual delineation, providing an efficient tool for simultaneous identification of 5 pairs of CNs without expert-defined ROIs. This approach significantly reduces operator-dependent bias and labor costs. Key observations are detailed below.

The proposed CN atlas successfully reconstructed all 5 pairs of CNs (II, III, V, VII/VIII) in individual subjects and achieved a 100\% identification rate for manually verifiable CNs across the cohort. Quantitative evaluation revealed strong spatial agreement between automated and manual selection methods \cite{he2021comparison,2020Anatomical,epprecht2024facial,huang2025uni}, with mean CN spatial overlap scores of 0.7448 (HCP subjects) and 0.7827 (MDM subjects)-exceeding the 0.72 threshold recommended by Cousineau et al. \cite{cousineau2017test}. Visual assessment confirmed highly similar fiber trajectories with robust spatial correspondence (Fig.~\ref{fig:3}). Importantly, these overlap scores are comparable to or exceed those reported for state-of-the-art single-pair CN atlases applied to individual nerves, such as CN II \cite{zeng2023automated}, CN III \cite{huang2022automatic}, CN V \cite{2020Creation}, and CN VII/VIII \cite{zeng2021automated}. This demonstrates that our multi-nerve approach achieves similar fidelity without sacrificing accuracy. We further demonstrated the method's robustness and reliability for CN identification using a dataset of 20 MDM subjects and two PA patients.

Unlike previous single-pair CN atlases, our proposed atlas aims to address these following challenges. Firstly, while effective for individual nerves, single-pair CN atlases struggle to provide a complete description of the multiple nerve pairs often involved around skull base tumors  \cite{sultana2017mri,jacquesson2025fiber,behan2017comparison}. To address this gap, we developed a comprehensive atlas specifically designed for the simultaneous mapping of five CN pairs.  Secondly, confirming prior observations on the sensitivity of CN tractography to parameter choice \cite{zolal2016comparison,yang2017visualization,jacquesson2019overcoming}, uniform parameters proved ineffective across diverse CNs. Building on this understanding, we systematically derived and implemented CN pair-specific multi-parametric tracking protocols within our atlas framework. Thirdly, structural variations between CNs lead to heterogeneous streamline distances, confounding standard fiber clustering methods \cite{li2024tractography,vazquez2020ffclust}. To overcome this, we introduced a novel multi-stage clustering strategy, explicitly designed to handle distance variability and minimize anatomically implausible streamlines specific to the skull base environment. 

We establish the anatomical validity of the proposed CN atlas in recognizing 5 pairs of CNs through multiple lines of evidence. First, the automatically identified CNs show high comparability to manual CN selection results, with visually similar fiber trajectories and strong spatial overlap (Fig.~\ref{fig:3}). Second, the anatomical validity of automatically recognized CNs was confirmed in patient data. As shown in Fig.~\ref{fig:4}, when a tumor compressed the optic nerve (CN II) along its pathway between the optic chiasm and lateral geniculate nucleus, our method successfully reconstructed the surrounding nerve fibers. This capability is critical for preoperative tumor resection planning. Finally, intraoperative validation in a CP patient confirmed that automatically identified nerves maintained clear spatial relationships with tumor locations.

The proposed CN atlas has significant applications in both clinical practice and scientific research. First, automatic CN identification addresses a critical need in skull base surgery \cite{huang2025uni,hu2024preoperative,upreti2024advancements}. While ROI-based tractography methods \cite{he2021comparison,2020Anatomical,epprecht2024facial,huang2025uni,golby1998trigeminal,jacquesson2019probabilistic} from diffusion MRI have been extensively studied, they require three labor-intensive stages: pre-processing, fiber tracking, and post-tracking processing. The post-tracking phase particularly demands manual ROI/ROA delineation and erroneous tract removal, requiring substantial time and specialized anatomical expertise. Our automated approach completes the entire process from seeding to final results in under 20 minutes, representing a substantial reduction compared to the more than 2 hours typically required for manual ROI-based CN tractography \cite{2020Anatomical} or the sequential processing needed when applying multiple single-pair atlases. Second, preoperative CN fiber reconstruction and visualization of spatial relationships with surrounding tissues are crucial for skull base surgery planning. Our method also shows potential for studying neuropathologies involving neurovascular conflicts, such as trigeminal neuralgia \cite{wood2020trigeminal} and facial spasms \cite{shapey2023diffusion}, by providing individualized neural structure identification to confirm such conflicts.

For the dataset used in this study, seed points and masks in MNI space were co-registered to individual subjects, with fiber tractography for each CN pair performed using the UKF method \cite{malcolm2010filtered}. Specific tracking parameters are detailed in Section III-E. Notably, tractography reconstruction quality directly influences clustering outcomes, as these are derived from streamline similarity assessments. While our methodology employs the UKF approach, alternative tractography techniques (deterministic, probabilistic \cite{dyrby2007validation}, and PTT \cite{aydogan2020parallel}) could be implemented for clinical data analysis, provided they generate anatomically plausible streamlines. Our clustering process can automatically identify CNs as long as the generated streamlines are anatomically correct.

Potential limitations of the present study, including suggested future work to address limitations, are as follows. First, in this study, we created the CN atlas using UKF tractography because it has been demonstrated to be effective in tracking CNs. However, the current optimal tracing parameters only apply to the specific single-pair CN \cite{he2021comparison,2020Anatomical,epprecht2024facial,huang2025uni}, and an interesting future work could include a comprehensive comparison to investigate the differences of the CNs identified from different tractography methods. Second, accurate imaging within the intracranial brainstem is beyond the reach of all current diffusion MRI tractography methods, and the course of CNs within the brainstem is less clear \cite{xie2023cntseg,yang2017visualization}. Therefore, this study focuses primarily on the cisternal segments of CNs, while the intracranial trajectories were reconstructed through ROI-based streamline filtering. This approach may present limitations in characterising decussating fibers within the brainstem, such as potential contralateral crossing of oculomotor nerve pathways. Future investigations that employ advanced high-resolution imaging techniques \cite{wang2024diff5t} will be essential for achieving precise visualisation and reconstruction of challenging brainstem-contained pathways of CNs. Thirdly, while our atlas currently targets five critical CN pairs, extending coverage to the remaining seven pairs (e.g., CN IV, VI, XI, XII) presents significant challenges due to factors like size, course, and proximity to other structures. Future work will explore combining our clustered atlas framework with advanced artificial intelligence algorithms \cite{li2024tractography,gao2025fine,chen2025tractgraphformer} to enable the automatic reconstruction of these additional CNs.

\section{Conclusion}\label{sec:Conclusion}
In this paper, we present the first comprehensive diffusion tractography atlas for automated mapping of the pathways in the human brain of CNs, addressing the challenges posed by the unique anatomical structures of CNs and the complexity of the skull base environment. By employing multi-parametric fiber tractography and an innovative multi-stage fiber clustering strategy, the proposed atlas successfully identifies 8 fiber bundles associated with 5 pairs of CNs with high anatomical accuracy. Quantitative and visual experiments using dMRI data from the HCP datasets, MDM datasets, two PA patients and a CP patient demonstrate that the automated mapping results are highly comparable to expert manual identification. 
\section{Data and Code Availability}
Publicly available datasets were used in this study: HCP dataset and the MDM dataset. The dMRI datasets are online available. The PA patients and CP patient are acquired at Xuanwu Hospital Capital Medical University. All data used in this study were simulated and did not require an ethical statement. The implementation code and the atlas data are publicly available on~\textit{https://github.com/IPIS-XieLei/CNsAtlas}, with detailed usage instructions provided. We hope this resource will enhance standardization in diffusion MRI analysis and welcome collaborations to continue the precision of this atlas and extend the atlas to other CNs.
%%Harvard
\bibliographystyle{IEEEtran}
\bibliography{IEEEabrv,myreference}
\end{document}